\documentclass{bmvc2k}

\usepackage[table]{xcolor}
\usepackage{multirow, tabularx}
\usepackage{booktabs}
\usepackage{adjustbox}

\title{Enhancement of SSD by concatenating feature maps for object detection 
}

\addauthor{Jisoo Jeong}{soo3553@snu.ac.kr}{1}
\addauthor{Hyojin Park}{wolfrun@snu.ac.kr}{1}
\addauthor{Nojun Kwak}{nojunk@snu.ac.kr}{1}

\addinstitution{
 Department of Transdisciplinary Studies \\ Seoul National University, Korea
}

\runninghead{J. Jeong, H. Park and N. Kwak}{Under Review in BMVC 2017}

\begin{document}
\maketitle

\begin{abstract}
We propose an object detection method that improves the accuracy of the conventional SSD (Single Shot Multibox Detector), 
which is one of the top object detection algorithms in both aspects of accuracy and speed. 
The performance of a deep network is known to be improved as the number of feature maps increases.
However, it is difficult to improve the performance by simply raising the number of feature maps. In this paper, we propose and analyze how to use feature maps effectively to improve the performance of the conventional SSD. 
The enhanced performance was obtained by changing the structure close to the classifier network, rather than growing layers close to the input data, e.g., by replacing VGGNet with ResNet. 
The proposed network is suitable for sharing the weights in the classifier networks, by which property, the training can be faster with better generalization power.
For the Pascal VOC 2007 test set trained with VOC 2007 and VOC 2012 training sets, the proposed network with the input size of $300 \times 300$ achieved 78.5\% mAP (mean average precision) at the speed of 35.0 FPS (frame per second), while the network with a $512 \times 512$ sized input achieved 80.8\% mAP at 16.6 FPS using Nvidia Titan X GPU. The proposed network shows state-of-the-art mAP, which is better than those of the conventional SSD, YOLO, Faster-RCNN and RFCN. Also, it is faster than Faster-RCNN and RFCN. 

\end{abstract}

\section{Introduction}
\label{sec:intro}

Object detection is one of the main areas of researches in computer vision. In recent years, convolutional neural networks (CNNs) have been applied to object detection algorithms in various ways, improving the accuracy and speed of object detection \cite{ren2015faster,yang2016exploit,redmon2016you,redmon2016yolo9000,liu2016ssd}. 

\begin{figure}[tb]
\begin{center}$
\begin{array}{c}
\includegraphics[width = 0.9\linewidth]{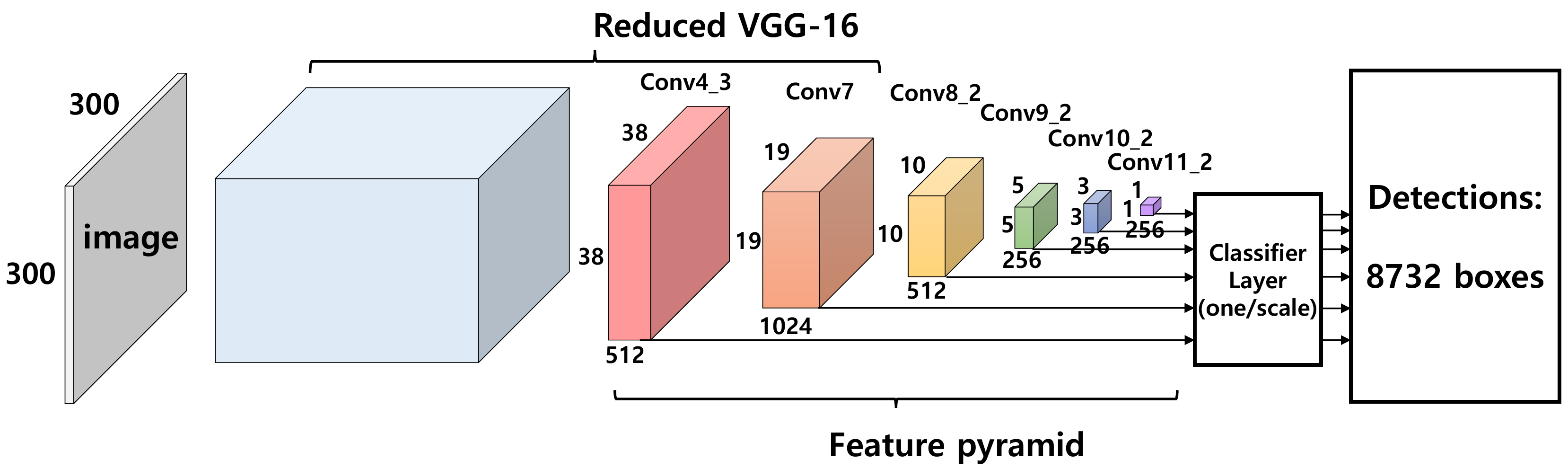}\\
\end{array}$
\end{center}
\caption{Overall structure of the conventional SSD. The layers from Conv4-3 to Conv11-2 which are used as the input to the classifier network is denoted as feature pyramid. Each layer in the feature pyramid is responsible for detecting objects in the corresponding size. 
}
\label{fig:Conventional SSD}
\end{figure}

Among various object detection methods, SSD \cite{liu2016ssd} is relatively fast and robust to scale variations because it makes use of multiple convolution layers for object detection. Although the conventional SSD performs good in both the speed and detection accuracy, it has a couple of points to be supplemented. First, as shown in Fig.~\ref{fig:Conventional SSD}, which shows the overall structure of conventional SSD, each layer in the feature pyramid\footnote{The term \textit{feature pyramid} is used to denote the set of layers that are directly used as an input to the classifier network as shown in Fig.~\ref{fig:Conventional SSD}.} is used independently as an input to the classifier network. 
Thus, the same object can be detected in multiple scales. 
%
Consider a certain position of a feature map in a lower layer (say, Conv4-3) is activated. This information can affect entire scales up to the the last layer (Conv11-2), which means that the relevant positions in the higher layers have a good chance to be also activated.
%

%
%
However, SSD does not consider the relationships between the different scales because it looks at only one layer for each scale. For example, in Fig.\ref{fig:ssdvspro}(a), SSD finds various scale boxes for one object.

Second, SSD has the limitation that small objects are not detected well. This is not the problem only for SSD but the problem for most object detection algorithms.
%
%
To solve this problem, there have been various attempts  such as replacing the base network with more powerful one, e.g., replacing VGGNet with ResNet
\cite{li2016r,fu2017dssd} or increasing the number of channels in a layer \cite{lawrence1998size}. Fig.\ref{fig:ssdvspro}(b) shows that SSD has a limitation in detecting small objects. Especially, in the two figures, persons on the boat and small cows are not detected, respectively.

In this paper, we tackle these problems as follows.
%
First, the classifier network is implemented considering the relationship between layers in the feature pyramid. Second, the number of channels (or feature maps) in a layer is increased efficiently. More specifically, only the layers in the feature pyramid are allowed to have increased number of feature maps instead of increasing the number of layers in the base network. 
%
The proposed network is suitable for sharing weights in the classifier networks for different scales, resulting in a single classifier network. This enables faster training speed with advanced generalization performance. 
Furthermore, this property of single classifier network is very useful in a small database. In the conventional SSD, if there is no object at a certain size, the classifier network of that size cannot learn anything. However, if a single classifier network is used, it can get information about the object from the training examples in different scales.

Using the proposed architecture, 
our version of SSD can prevent detecting multiple boxes for one object as shown in Fig.\ref{fig:ssdvspro}(c). In addition, the number of channels can be efficiently increased to detect small objects as shown in Fig.\ref{fig:ssdvspro}(d). 
The proposed method shows state-of-the-art mAP (mean average precision) with a slightly degraded speed compared to the conventional SSD.

The paper is organized as follows. In Section \ref{sec:related}, we briefly review the related works in the area of object detection, and then propose a different version of SSD, which is called as the \textit{rainbow SSD}, in Section \ref{sec:method}. The experiments and the evaluation of the proposed algorithm are presented in Section \ref{sec:experiment}. 
In Section \ref{sec:discussion}, a discussion with some exemplary results of the proposed method is made. Finally, the paper is concluded in Section \ref{sec:conclusion}.

\begin{figure}[t]
\begin{center}$
\begin{array}{cccc}
\includegraphics[width=2.8cm]{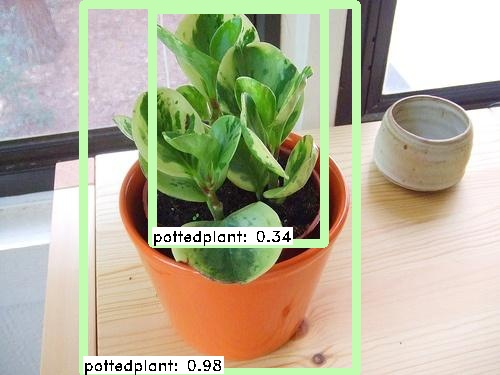}&
\includegraphics[width=2.8cm]{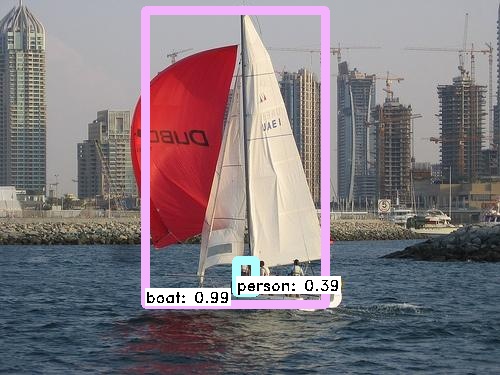}&
\includegraphics[width=2.8cm]{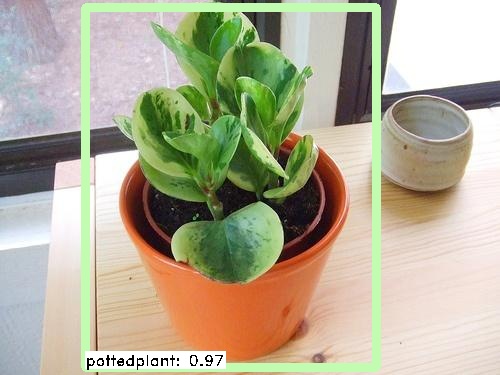}&
\includegraphics[width=2.8cm]{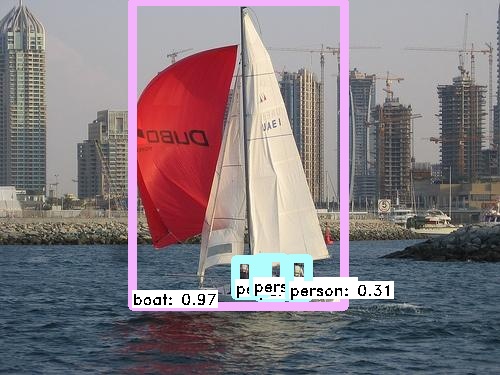}\\
\includegraphics[width=2.8cm]{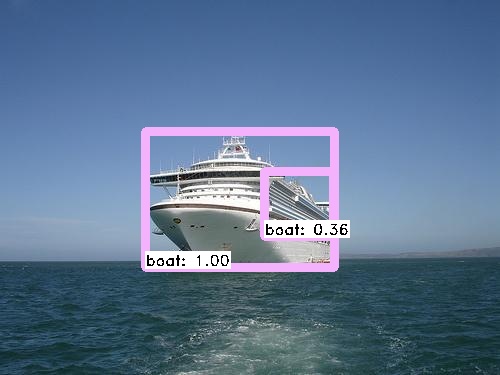}&
\includegraphics[width=2.8cm]{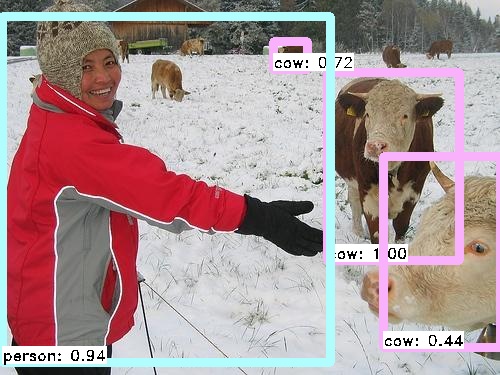}&
\includegraphics[width=2.8cm]{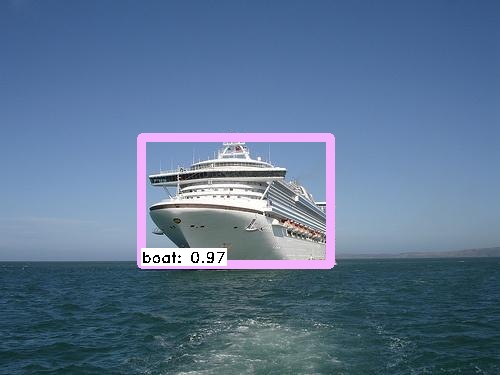}&
\includegraphics[width=2.8cm]{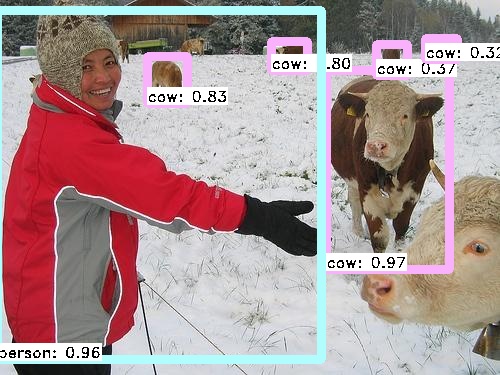}\\
(a)&(b)&(c)&(d)
\end{array}$
\end{center}
\caption{Conventional SSD vs. the proposed Rainbow SSD (R-SSD).  Boxes with objectness score of 0.3 or higher is drawn: (a) SSD with two boxes for one object; (b) SSD for small objects; (c) R-SSD with one box for one object; (d) R-SSD for small objects}
\label{fig:ssdvspro}
\end{figure}

\section{Related Works}
\label{sec:related}

A wide variety of methods using deep learning have been applied to the problem of object detection and it continues to show performance improvements. In the earlier works pioneered by R-CNN (region-based CNN) \cite{girshick2014rich}, the candidate region was proposed through a separate algorithms such as selective search \cite{uijlings2013selective} or Edge boxes \cite{zitnick2014edge} and the classification was performed with deep learning. 

Although R-CNN improved the accuracy using deep learning, speed was still a problem, and end-to-end learning was impossible. The region proposal network (RPN) was first proposed in faster R-CNN, which improved the speed of the object detector significantly and was able to learn end-to-end \cite{ren2015faster}. 

YOLO (you only look once) greatly improved the speed by dividing a single image into multiple grids and simultaneously performing localization and classification in each grid \cite{redmon2016you}. While YOLO performed object detection by concentrating only on speed, an enhanced version of YOLO, which is denoted as YOLO2, removed the fully connected layers and used anchor boxes to improve both the speed and the accuracy \cite{redmon2016yolo9000}. 

On the other hand, SSD creates bounding box candidates at a given position and scale and obtains their actual bounding box and score for each class \cite{liu2016ssd}. Recently, to improve the accuracy of SSD, especially for small object, DSSD (deconvolutional SSD) that uses a large scale context for the feature pyramid was proposed \cite{fu2017dssd}. DSSD applied a deconvolution module to the feature pyramid and used ResNet instead of VGGNet. DSSD succeeded in raising accuracy at the expense of speed.


Besides the fields of object detection, the fields of segmentation have also been much developed by the application of deep neural networks. Many of them use pixel-based object classification in combination with upsampling or deconvolution to obtain the segmentation results in the same size as the input image \cite{noh2015learning,yu2015multi}. In addition, features are engineered in various ways by concatenation, element-wise addition or element-wise product with bypass, to obtain improved performance. This paper is inspired by the fact that we get improved abstract representation of the original image from features in different scales \cite{hariharan2015hypercolumns,bansal2016pixelnet}. 

\section{Method}
\label{sec:method}

As mentioned above, our strategy of improving the accuracy of SSD is to let the classifier network fully utilize the relationship between the layers in the feature pyramid
without changing the base network that is closely located to the input data. 
In addition, it also increases the number of channels in the feature pyramid efficiently.

\begin{figure*}[t]
\begin{center}
\includegraphics[width = 0.8\linewidth]{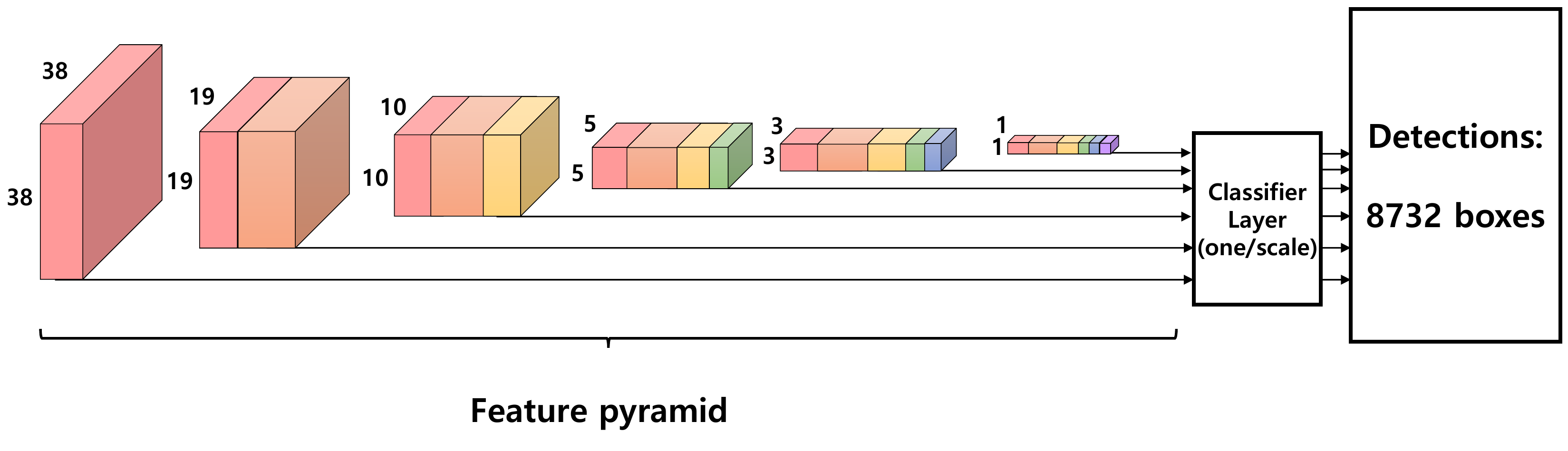}\\
(a) pooling\\
\includegraphics[width = 0.8\linewidth]{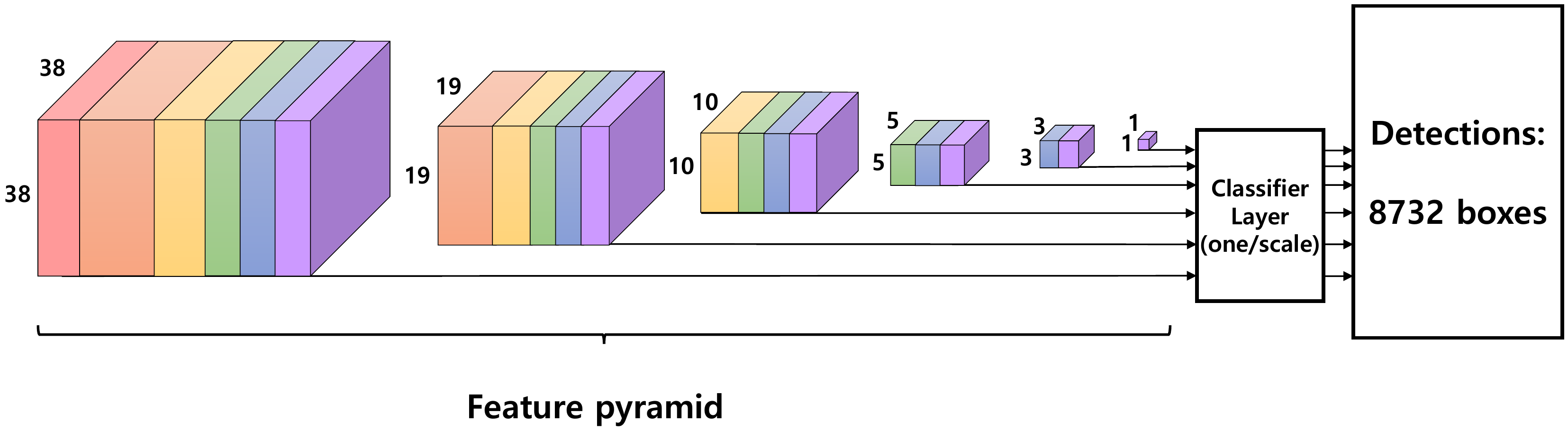}\\
(b) deconvolution\\
\includegraphics[width = 0.8\linewidth]{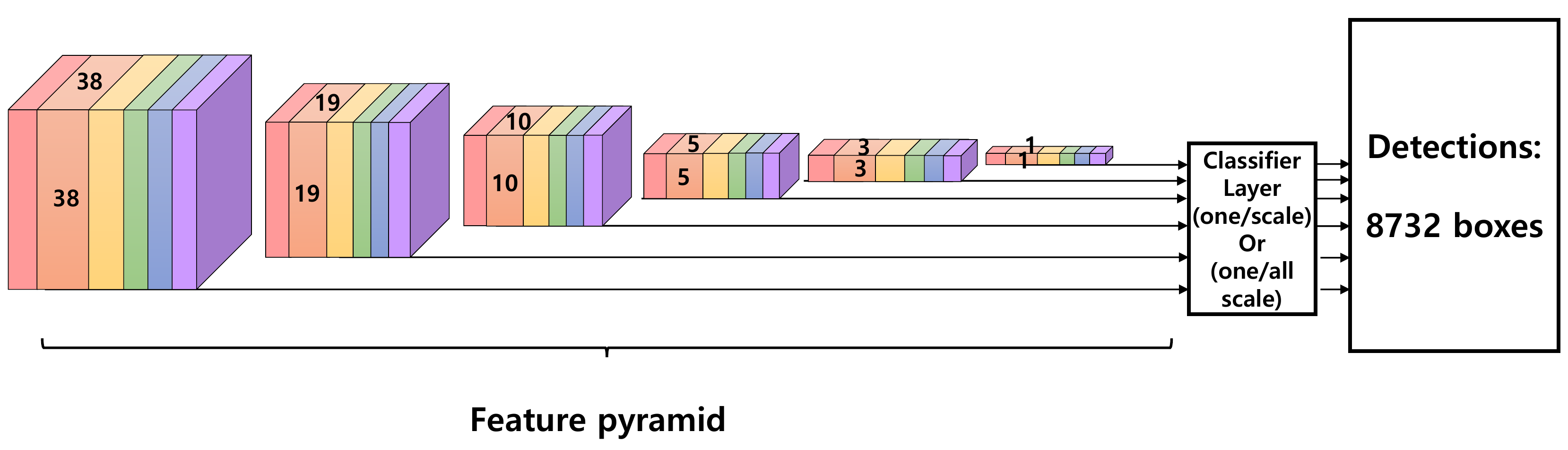}\\
(c) both pooling and deconvolution (Rainbow concatenation)
\end{center}
\caption{Proposed methods of feature concatenation: (a) concatenation through pooling (b) concatenation through  deconvolution; (c) rainbow concatenation through both pooling and concatenation. (best viewed in color. See Fig.~\ref{fig:Conventional SSD} for comparison.)}
\label{fig:method}
\end{figure*}

Figure \ref{fig:method} shows several ways of increasing the number of feature maps in different layers for the classifier networks to utilize the relationship between layers in the feature pyramid. To enable this, in Fig.~\ref{fig:method}(a), feature maps in the lower layers are concatenated to those of the upper layers through pooling. In this way, the classifier networks with large receptive fields can have enriched representation power for object detection.
On the other hand, Fig.~\ref{fig:method}(b) shows the method of concatenating the feature maps of the upper layers to the lower layer features through deconvolution or upsampling. 
Fig.~\ref{fig:method}(c) shows the feature map concatenation method that utilize both the lower layer pooling and the upper layer deconvolution. 

One thing to note is that before concatenating feature maps, a normalization step is inevitable. This is because the feature values in different layers are quite different in scale. Here, batch normalization \cite{ioffe2015batch,liu2015parsenet} is applied for each filter before concatenation. 
%

All of the above methods have the advantage that end-to-end learning is possible. More details about each are described below.



\subsection{Concatenation through pooling or deconvolution}
\label{subsec:pooling}

In the structure of SSD, generally, the numbers of channels in the lower layers are larger than those in the upper layer. To make explicit relationship between feature pyramid and to increase the number of channels effectively, we concatenate feature maps of the upper layers through pooling or concatenate feature maps of the lower layers through deconvolution. Unlike DSSD \cite{fu2017dssd}, which uses deconvolution module consisting of 3 convolution layer, 1 deconvolution layer, 3 batch normalization layer, 2 Relu and elementwise product, our model of concatenation through deconvolution performs only deconvolution with batch normalization and does not need  elementwise product. 

The advantage of these structure is that object detection can be performed with information from the other layers. On the other hand, the disadvantage is that information flows unidirectional and the classifier network cannot utilize other directional information.


\subsection{Rainbow concatenation}
\label{subsec:R-SSD}

As shown in Fig.~\ref{fig:method}(c), in the rainbow concatenation, pooling and deconvolution are performed simultaneously to create feature maps with an explicit relationship between different layers. After pooling or deconvolving features in every layers to the same size, we concatenate them. Using this concatenated features, detection is performed considering all the cases where the size of the object is smaller or larger than the specific scale.
That is, it can have additional information about the object larger than or smaller than the object. 
Therefore, it is expected that the object in a specific size is likely to be detected only in an appropriate layer in the feature pyramid as shown in Fig.~\ref{fig:ssdvspro}(c). 

In addition, the low-layer features that has been with limited representation power are enriched by the concatenation of higher-layer features, resulting in good representation power for small object detection as in DSSD \cite{fu2017dssd} without much computational overhead.


By rainbow concatenation, each layer of feature pyramid contains 2,816 feature maps (concatenation of 512, 1024, 512, 256, 256, and 256 channels) in total. Because each layer in the feature pyramid now has the same number of feature maps, weights can be shared for different classifier networks in different layers. Each classifier network of the conventional SSD checks 4 or 6 default boxes and the proposed rainbow SSD can unify default boxes in different layers with weight sharing. As shown in Table \ref{tab:numberofbox}, conventional SSD makes a 8,732 total boxes. On the other hand, in the shared classifier with 4 default boxes for each layer, the number becomes 7,760, likewise, for the shared classifier with 6 default boxes, it becomes 11,640 in total.

\begin{table}[tbp]
  \centering
  \small
    \begin{tabular}{|@{\,}c@{\,}|@{\,}c@{\,}@{\,}c@{\,}@{\,}c@{\,}@{\,}c@{\,}@{\,}c@{\,}@{\,}c@{\,}|@{\,}c@{\,}|}
    \hline
          \# of box positions & $38 \times 38$ & $19 \times 19$ & $10 \times 10$ & $5 \times 5$   & $3 \times 3$   & $1 \times $   & total boxes \\
          \hline
          \hline
    conventional SSD & 4     & 6     & 6     & 6     & 4     & 4     & 8732 \\
    R-SSD with one classifier (4 boxes) & 4     & 4     & 4     & 4     & 4     & 4     & 7760 \\
    R-SSD with one classifier (6 boxes) & 6     & 6     & 6     & 6     & 6     & 6     & 11640 \\
    \hline
    \end{tabular}
  \caption{The number of boxes for each classifier network and the number of total boxes}
  \label{tab:numberofbox}
\end{table}%



\subsection{Increasing number of channels}
\label{subsec:increasing}

\begin{table*}[tbp]
  \centering
  \small
    \begin{tabular}{|l|cccr|cc|cc|cc|}
    \hline
          & \multicolumn{4}{c|}{conv4}    & \multicolumn{2}{c|}{conv7} & \multicolumn{2}{c|}{conv8} & \multicolumn{2}{c|}{conv9 - conv11} \\
          
\cline{2-11}          & \multicolumn{3}{c}{1 - 3} & 4     & 1     & 2     & 1     & 2     & 1     & 2 \\
    \hline
    \hline
    SSD   & \multicolumn{3}{c}{512} &       & 1024  &       & 256   & 512   & 128   & 256 \\
    I-SSD & \multicolumn{3}{c}{512} & 2048  & 1024  & 2048  & 1024  & 2048  & 1024  & 2048 \\
    \hline
    \end{tabular}%
  \caption{Number of channels in conventional SSD and I-SSD}
  \label{tab:increasedchannel}%
\end{table*}%

It is known that the larger the number of channels, the better the performance becomes. To demonstrate how efficient our method is, we present a comparison model, for which we simply change the number of channel in the original base network as shown in Table \ref{tab:increasedchannel}. As in SSD, reduced VGG-16 pre-trained model is used as the base network. The number of channels in each convolution layer are set to be 2 to 8 times larger than the original network. From now on, this comparison network will be denoted as I-SSD, which stands for increased-channel SSD.

\section{Experiments}
\label{sec:experiment}

\begin{table}[tb]
  \centering
  \footnotesize
    \begin{tabular}{|@{\,}l@{\,}|@{\,}c@{\,}|@{\,}c@{\,}|@{\,}r@{\,}|@{\,}r@{\,}|@{\,}r@{\,}|}
    \hline
          & \multicolumn{1}{l|}{Input} & Train & \multicolumn{1}{l|}{Test} & \multicolumn{1}{l|}{mAP} & \multicolumn{1}{l|}{FPS} \\
    \hline
    \hline
    YOLO\cite{redmon2016you} & 448   & VOC2007+2012 & 2007  & 63.4  & 45 \\
    YOLOv2\cite{redmon2016yolo9000}  & 416 & VOC2007+2012 & 2007   & 76.8  & 67 \\
    YOLOv2 544x544\cite{redmon2016yolo9000}  & 544 & VOC2007+2012 & 2007   & 78.6  & 40 \\
    Faster R-CNN\cite{ren2015faster}  &       & VOC2007+2012 & 2007  & 73.2  & 5 \\
    R-FCN (ResNet-101)\cite{li2016r} &  & VOC2007+2012 & 2007   & \textbf{80.5} & 5.9 \\
    \hline
    SSD*\cite{liu2016ssd}  & 300  & VOC2007+2012 & 2007   & \textbf{77.7}  & 61.1 \\
    DSSD (ResNet-101)\cite{fu2017dssd} & 321 & VOC2007+2012 & 2007 & 78.6 & 9.5\\
    ISSD*  & 300 & VOC2007+2012 & 2007   & 78.1  & 26.9 \\
    ours (SSD pooling)*  & 300 & VOC2007+2012 & 2007   & 77.1  & 48.3 \\
    ours (SSD deconvolution)*  & 300 & VOC2007+2012 & 2007   & 77.3  &  39.9 \\
    \hline
    ours (R-SSD)*  & 300 & VOC2007+2012 & 2007   & \textbf{78.5} & 35.0 \\
    ours (R-SSD one classifier (4 boxes))*  & 300 & VOC2007+2012 & 2007   & \textbf{76.2} & 34.8\\
    ours (R-SSD one classifier (6 boxes))*  & 300 & VOC2007+2012 & 2007   & \textbf{77.0} & 35.4\\
    
    \hline
    SSD*\cite{liu2016ssd}  & 512 & VOC2007+2012 & 2007   & 79.8  & 25.2 \\
    DSSD (ResNet-101)\cite{fu2017dssd} & 513 & VOC2007+2012 & 2007 & 81.5 & 5.5\\
    ours (R-SSD)*  & 512 & VOC2007+2012 & 2007   & \textbf{80.8} & 16.6 \\
    \hline
    \end{tabular}%
    \caption{VOC2007+2012 training and VOC 2007 test result (* is tested by ourselves)}
    \label{tab:VOC2007}%
\end{table}%

In order to evaluate the performance of the proposed algorithm, we tested various versions of SSD for PASCAL VOC2007 dataset\cite{everingham2010pascal}. 
The VOC dataset consists of 20 object classes
with the annotated ground truth location and the corresponding class information for each image.




We trained our model with VOC2007 and VOC2012 `trainval' datasets. The SSDs with $300\times 300$ input were trained with a batch size of 8, and the learning rate was reduced from $10^{-3}$ to $10^{-6}$ by $10^{-1}$. With each learning rate, we trained 80K, 20K, 20K, and 20K iterations respectively. Thus the total number of iteration for each model was 140K. The $512 \times 512$ input models were performed with a batch size of 4 and the learning rate was equal to that for $300 \times 300$ models. 
In the case of speed, it is measured by using the forward path of the network with a batch size of 1. The experiments were done with cuDNN v5.1 using CAFFE time function. Therefore, if the detection time is measured from the pre-processing (resizing image and so on), it may take longer. The experimental results are shown in Table \ref{tab:VOC2007}. In the table, the performances of YOLO \cite{redmon2016you}, YOLOv2 \cite{redmon2016yolo9000}, Faster R-CNN \cite{ren2015faster}, R-FCN \cite{li2016r}, and DSSD \cite{fu2017dssd} were obtained from their homepage \footnote{YOLO and YOLOv2 : http://pjreddie.com/darknet} or the respective paper. To see the performance of various feature augmentation methods, we performed experiments using features concatenated through pooling (SSD pooling) and deconvolution (SSD deconvolution) as described in Section \ref{subsec:pooling}. We also tested ISSD described in Section \ref{subsec:increasing} for comparison. Three types of R-SSD was tested. The first one utilizes separate classifier networks for different scales and the rest two use a common classifier with 4 or 6 default boxes for a scale as described in Section \ref{subsec:R-SSD}. The conventional SSD was also trained and tested by ourselves.

%

\textbf{ISSD:} 
For the 300 input model, we experimented ISSD by increasing the number of the channels from 2 to 8 times for different layers as Table \ref{subsec:increasing}. As a result, there is a 0.4\% mAP improvement in accuracy with 78.1\% mAP compared to conventional SSD. However, due to the increased number of channels, the speed drops to 26.9 FPS  

\textbf{Concatenation through pooling or deconvolution:} 
For the 300 input model, the concatenation through pooling and concatenation through deconvolution result in mAP of 77.1\%  and 77.3\% respectively which is a degraded performance than that of conventional SSD by 0.6\% and 0.4\%. In addition, due to the increased complexity, the speed was also degraded to 48.3 FPS and 39.9 FPS, respectively. 

\textbf{R-SSD: }
For the 300 input model, there is a 0.8\% improvement in accuracy with 78.5\% mAP compared to conventional SSD. However, due to the increased computational complexity, the speed drops to 35.0 FPS. For the 512 input model, it results in mAP of 80.8\% which is 1\% better than conventional SSD. However its speed drops to 16.6 FPS. In particular, comparing the two SSD 512 models, precision increases 2.9\% at recall value 0.8 and 8.2\% at recall of 0.9.
In the case of single classifier model with 300 input model, it has a 76.2\% and 77.0\% mAP when they use four and six default boxes, respectively. 





\section{Discussion}
\label{sec:discussion}

\subsection{New evaluation method}

\begin{figure}[t]
\begin{center}
\includegraphics[width=0.5\linewidth]{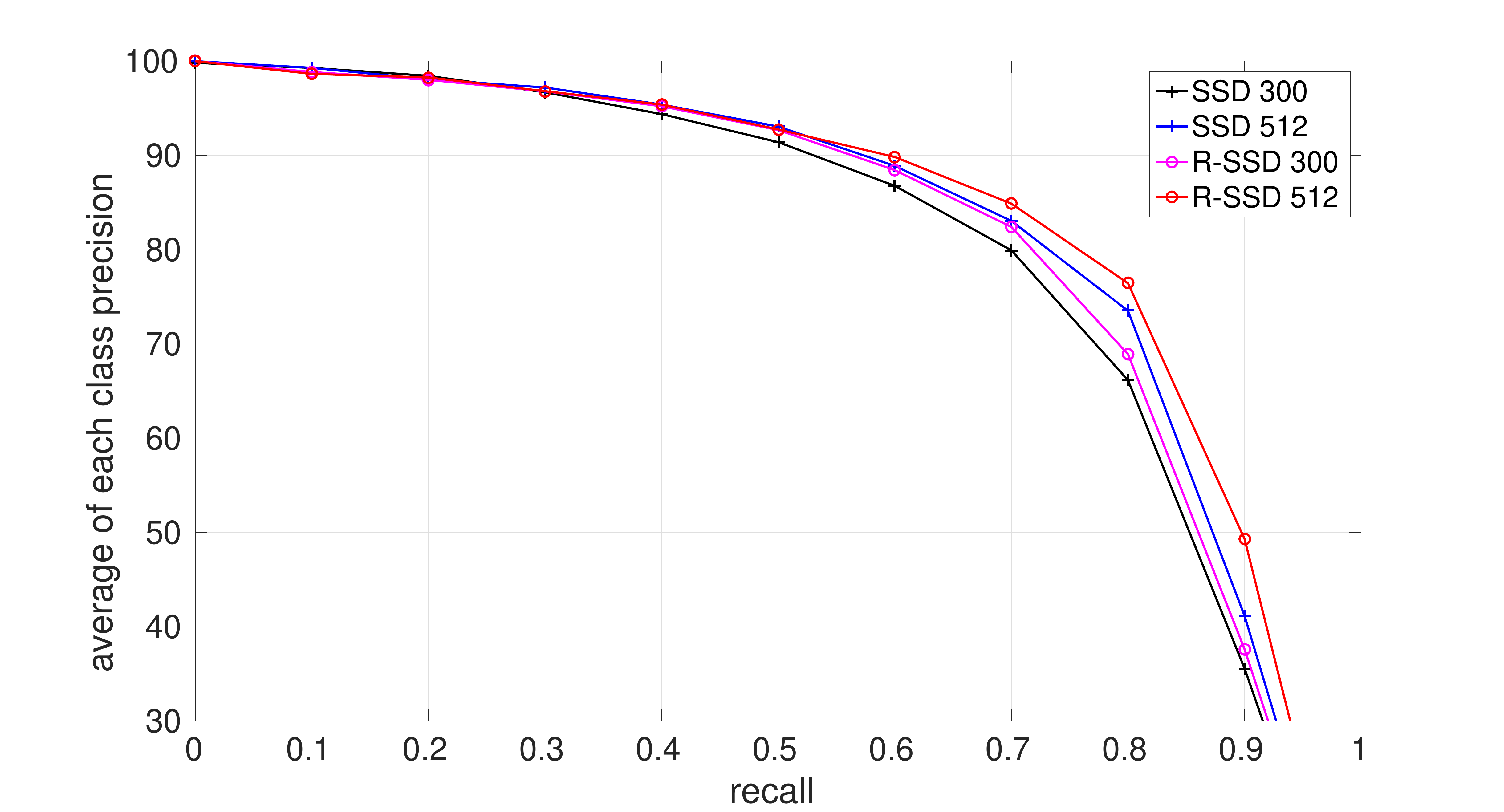}
\end{center}
\caption{Recall vs. average of each class precision graph in PASCAL VOC 2007 test}
\label{fig:recall}
\end{figure}

\begin{table}[tbp]
  \centering
  \begin{adjustbox}{width=\linewidth}
    \begin{tabular}{|l|r|l|rrrrrr|r|}
    
    \hline
    \multirow{2}[4]{*}{} & \multicolumn{1}{c|}{\multirow{2}[4]{*}{Input}} & \multicolumn{1}{c|}{\multirow{2}[4]{*}{Train}} & \multicolumn{6}{c|}{Recall}                   &
    \multicolumn{1}{c|}{\multirow{2}[4]{*}{mAP@0.7+}} \\
\cline{4-9}          &       &       & 0.5   & 0.6   & 0.7   & 0.8   & 0.9   & 1     &   \\
    \hline
    \hline
    SSD   & 300   & VOC2007+2012 & 91.4  & 86.8  & 80.0    & 66.2  & 35.6  & 0     & 45.5 \\
\cline{1-3}\cline{10-10}    R-SSD (ours)   & 300   & VOC2007+2012 & 92.7  & 88.4  & 82.4  & 68.9  & 37.6  & 0     & 47.2 \\
\hline   
SSD   & 512   & VOC2007+2012 & 93.1  & 88.9  & 83.0    & 73.5  & 41.1  & 0     & 49.4 \\
\cline{1-3}\cline{10-10}    R-SSD (ours) & 512   & VOC2007+2012 & 92.8  & 89.8  & 84.9  & 76.4  & 49.3  & 0     & 52.7 \\
\hline
    \end{tabular}%
    \end{adjustbox}
    \caption{Average of precisions over all the classes for fixed recall and mAP@0.7+ (VOC evaluation Toolkit was used)}
  \label{tab:mAP0.7}%
\end{table}%


The most commonly used evaluation technique for object detection is mAP. AP (Average Precision) is a concept of integrating precision as recall is varied from 0 to 1 and mAP is defined as the average of AP for all the object classes. 

In Figure~\ref{fig:recall}, we show the recall vs. average precision graph for PASCAL VOC 2007 test data. Note that average precision (vertical axis) here is averaged value for all the object classes. In the figure, we can see that different versions of SSD (SSD 300, SSD 512, R-SSD 300, R-SSD 512) have almost the same average precision for small ($<0.5$) recall values. Due to this, even if the precisions for large recall values show significant difference between algorithms (around 14\% difference for SSD 300 and R-SSD 512), the difference in mAP is relatively small (around 3\% for SSD 300 and R-SSD 512). 
Considering the use case of most object detection algorithms such as for autonomous vehicles, the precision value measured at high ($> 0.7$) recall is more important than that measured at small recall value. Therefore, in Table \ref{tab:mAP0.7}, we show the mAP computed by averaging only the APs at recall of 0.7 or higher. In the table, we can see more clearly the effectiveness of R-SSD over SSD. At recall of 0.9, R-SSD (512) outperformed SSD (512) by more than 8\%. 
Note that the APs at recall of 1 is 0 for all the cases. This is due to the fact that we cannot recall all the objects in the test images regardless how we lower the score threshold.  



\subsection{Concatenation by pooling or deconvolution}

These two models are both inferior in accuracy and speed compared to conventional SSD, although they made explicit relationship between multiple layers and increased the number of channels. These two models need to perform more operations, therefore the speed can drop. As for accuracy, the reason can be conjectured that the layers sharing the same feature maps with other layers can be affected by the loss of other scales and do not fully focus on the scale. That is, they cannot learn properly on their scale. 

\subsection{Single classifier vs. Multiple classifiers}

Unlike the conventional SSD, because R-SSD have the similar feature maps for different layers only different in size, the classifier network can be shared. Here, the experiments with a single classifier network by unifying the number of channels in each scale of feature pyramid. As shown in the Table \ref{tab:VOC2007}, there is a difference in the number of boxes, but there is little difference in speed. In comparison, performance was 1.2 \% and 0.7 \% lower than that of conventional SSD. However, the advantage of a single classifier is that learning can be effective especially when there are significant imbalance between the numbers of training samples for different sizes. In this case, conventional SSD cannot train the classifier for a scale with small number of samples. However, in R-SSD, this problem is avoided because the classifier network is shared. Furthermore, single classifier is faster at the early stage of training. Therefore, even for a large dataset, R-SSD can be trained fast by training a single classifier in the early stage and at some point, the classifiers can be trained separately for different scales.

\subsection{Accuracy vs. Speed}

The conventional SSD is one of the top object detection algorithms in both aspects of accuracy and speed. For SSD300 or SSD512, it has 77.7 \% mAP and 79.8 \% mAP respectively and has 61.1 FPS and 25.2 FPS. Looking at the results of the ISSD for comparison with our algorithm, ISSD had a 0.4 \% accuracy gain, but the speed dropped to 26.9 FPS. In our experiments, R-SSD shows improved accuracy with a bit slow speed. Compared to the ISSD, R-SSD shows higher accuracy and faster speed. Moreover, it shows about 1\% mAP improvement over the conventional SSD. At a speed of 15 fps or higher, our R-SSD shows 80.8 \% mAP. Compared to R-FCN with similar accuracy, R-SSD is about three times faster.

\subsection{Performances for different scales}

Table \ref{tab:sizerecall} shows the recall of each object size \cite{lin2014microsoft}. 
Normally, the AP or AR (Average Recall) should be obtained, but, the VOC2007 test set has a total of 12,032 objects, of which 567 are small objects. In evaluating the performance for small objects, there are several classes with no object at all. Therefore, we integrate all the objects in measuring the recall. 
When the object size is small, R-SSD300 and R-SSD512 detect more number of objects than SSD300 and SSD512, respectively. It can be shown that R-SSD misses a few small objects. At medium size, R-SSD300 has even more recall than SSD512. 
Furthermore, when the object size is large, recall of all models show high value over 0.93. The difference of R-SSD300, SSD512, and R-SSD512 is less than 10 out of 7,641. 

\begin{table}[tbp]
  \centering
  \small
    \begin{tabular}{|c|c|c|c|}
    \hline
          & \multicolumn{3}{c|}{Recall (\# of detected objects / \# of total object)} \\
\cline{2-4}          & Small (area<$32^2$) & Medium ($32^2$<area<$96^2$) & Large ($96^2$<area) \\
    \hline
    \hline
    SSD300 & 0.3845 (218/567) & 0.7754 (2965/3824) & 0.9314 (7117/7641) \\
    ours(R-SSD300) & 0.4127 (234/567) & 0.8073 (3087/3824) & 0.9374 (7163/7641) \\
    SSD512 & 0.6526 (370/567) & 0.8023 (3068/3824) & 0.9361 (7153/7641) \\
    ours(R-SSD512) & 0.6949 (394/567) & 0.8248 (3154/3824) & 0.9365 (7156/7641) \\
    \hline
    \end{tabular}%
  \caption{Recall for objects in different size \cite{lin2014microsoft}. A box is declared as an object when the object score is higher than 0.1.}
  \label{tab:sizerecall}%
\end{table}%


\section{Conclusion}
\label{sec:conclusion}

In this paper, we presented a rainbow concatenation scheme that can efficiently solve the problems of the conventional SSD. The contribution of the paper is as follows. First, it creates a relationship between each scale of feature pyramid to prevent unnecessary detection such as multiple boxes in different scales for one object. Second, by efficiently increasing the number of feature maps of each layer in the feature pyramid, the accuracy is improved without much time overhead. Finally, the number of feature maps for different layers are matched so that a single classifier can be used for different scales. By using a single classifier, improvement on the generalization performance can be expected, and it can be effectively used for datasets with size imbalance or for small datasets. The proposed R-SSD was created considering both the accuracy and the speed simultaneously, and shows state-of-the-art mAP among the ones that have speed of more than 15 FPS.

\section{Additional Experiments}
\label{sec:add_exp}

To show that the proposed method performs well especially for a small training dataset, we trained the networks using the \textit{train dataset} in VOC2007 and compared the proposed methods with others. 
Also, we trained the networks using PASCAL VOC2012\cite{everingham2010pascal} datasets and show the detection results here.

\subsection{Small Train Dataset in VOC2007}
\label{sec:smalldata}

Table \ref{tab:smalldata} is experimental results of different networks that were trained using the \textit{train dataset} in VOC2007 which consists of a relatively small number of images (2,501 images in total). Each network was trained with these 2,501 images and the mAP was measured with VOC 2007 test dataset. The conventional SSD \cite{liu2016ssd} shows a mAP of  66.1\%, while R-SSD achieves 66.9\% which is 0.8\% better than that of SSD. Furthermore, R-SSD with one classifier achieves an even better mAP of 67.2\%. It shows that training a single classifier is better not only in generalization power resulting in a faster training speed but also in detection performance when the training dataset is small.

\begin{table}[htbp]
  \centering
  
    \begin{tabular}{|c@{\,}|@{\,}c@{\,}|@{\,}c@{\,}|@{\,}r@{\,}|}
    \hline
          & Input & pre-trained model & {mAP}  \\
    \hline
    \hline
    SSD \cite{liu2016ssd}   & 300   & reduced VGG-16 &  66.1 \\
    ours (R-SSD) & 300   & reduced VGG-16 &  66.9 \\
    ours (R-SSD one classifier (6 boxes)) & 300   & reduced VGG-16 &  67.2 \\
    \hline
    \end{tabular}%
    \caption{Results on VOC2007 test dataset trained with VOD2007 small train dataset (2,501 images)}
  \label{tab:smalldata}%
\end{table}%

\subsection{VOC2012}
\label{sec:voc2012}


\begin{table}[htbp]
\begin{tiny}
\renewcommand{\tabcolsep}{0.8mm}
  \centering
  

    \begin{tabular}{|@{\,}c@{\,}|@{\,}c@{\,}|@{\,}c@{\,}@{\,}|@{}c@{}@{}c@{}@{}c@{}@{}c@{}@{}c@{}@{}c@{}@{}c@{}@{}c@{}@{}c@{}@{}c@{}@{}c@{}@{}c@{}@{}c@{}@{}c@{}@{}c@{}@{}c@{}@{}c@{}@{}c@{}@{}c@{}@{}c@{}|}
    \hline
    Method &  Train & mAP   & \multicolumn{1}{c}{aero} & \multicolumn{1}{c}{bike} & \multicolumn{1}{c}{bird} & \multicolumn{1}{c}{boat} & \multicolumn{1}{c}{bottle} & \multicolumn{1}{c}{bus} & \multicolumn{1}{c}{car} & \multicolumn{1}{c}{cat} & \multicolumn{1}{c}{chair} & \multicolumn{1}{c}{cow} & \multicolumn{1}{c}{table} & \multicolumn{1}{c}{dog} & \multicolumn{1}{c}{horse} & \multicolumn{1}{c}{mbike} & \multicolumn{1}{c}{person} & \multicolumn{1}{c}{plant} & \multicolumn{1}{c}{sheep} & \multicolumn{1}{c}{sofa} & \multicolumn{1}{c}{train} & \multicolumn{1}{c|}{tv} \\
    \hline
    \hline
    YOLO \cite{redmon2016you}  & 07++12 & 57.9  & \multicolumn{1}{c}{77.0}    & \multicolumn{1}{c}{67.2} & \multicolumn{1}{c}{57.7} & \multicolumn{1}{c}{38.3} & \multicolumn{1}{c}{22.7} & \multicolumn{1}{c}{68.3} & \multicolumn{1}{c}{55.9} & \multicolumn{1}{c}{81.4} & \multicolumn{1}{c}{36.2} & \multicolumn{1}{c}{60.8} & \multicolumn{1}{c}{48.5} & \multicolumn{1}{c}{77.2} & \multicolumn{1}{c}{72.3} & \multicolumn{1}{c}{71.3} & \multicolumn{1}{c}{63.5} & \multicolumn{1}{c}{28.9} & \multicolumn{1}{c}{52.2} & \multicolumn{1}{c}{54.8} & \multicolumn{1}{c}{73.9} & \multicolumn{1}{c|}{50.8} \\
    YOLOv2 544 \cite{redmon2016yolo9000} & 07++12 & 73.4  & \multicolumn{1}{c}{86.3}  & \multicolumn{1}{c}{82.0} & \multicolumn{1}{c}{74.8} & \multicolumn{1}{c}{59.2} & \multicolumn{1}{c}{51.8} & \multicolumn{1}{c}{79.8} & \multicolumn{1}{c}{76.5} & \multicolumn{1}{c}{90.6} & \multicolumn{1}{c}{52.1} & \multicolumn{1}{c}{78.2} & \multicolumn{1}{c}{58.5} & \multicolumn{1}{c}{89.3} & \multicolumn{1}{c}{82.5} & \multicolumn{1}{c}{83.4} & \multicolumn{1}{c}{81.3} & \multicolumn{1}{c}{49.1} & \multicolumn{1}{c}{77.2} & \multicolumn{1}{c}{62.4} & \multicolumn{1}{c}{83.8} & \multicolumn{1}{c|}{68.7} \\
    \hline
    SSD300* \cite{liu2016ssd} & 07++12 & 75.6  &  \multicolumn{1}{c}{88.2}  & \multicolumn{1}{c}{82.9} & \multicolumn{1}{c}{74.0} & \multicolumn{1}{c}{61.1} & \multicolumn{1}{c}{47.3} & \multicolumn{1}{c}{82.9} & \multicolumn{1}{c}{78.9} & \multicolumn{1}{c}{91.6} & \multicolumn{1}{c}{57.8} & \multicolumn{1}{c}{80.2} & \multicolumn{1}{c}{63.9} & \multicolumn{1}{c}{89.3} & \multicolumn{1}{c}{85.5} & \multicolumn{1}{c}{85.9} & \multicolumn{1}{c}{82.2} & \multicolumn{1}{c}{49.7} & \multicolumn{1}{c}{78.8} & \multicolumn{1}{c}{73.1} & \multicolumn{1}{c}{86.6} & \multicolumn{1}{c|}{71.6} \\
    DSSD321 \cite{fu2017dssd} &  07++12 & 76.3  & \multicolumn{1}{c}{87.3}  & \multicolumn{1}{c}{83.3} & \multicolumn{1}{c}{75.4} & \multicolumn{1}{c}{64.6} & \multicolumn{1}{c}{46.8} & \multicolumn{1}{c}{82.7} & \multicolumn{1}{c}{76.5} & \multicolumn{1}{c}{92.9} & \multicolumn{1}{c}{59.5} & \multicolumn{1}{c}{78.3} & \multicolumn{1}{c}{64.3} & \multicolumn{1}{c}{91.5} & \multicolumn{1}{c}{86.6} & \multicolumn{1}{c}{86.6} & \multicolumn{1}{c}{82.1} & \multicolumn{1}{c}{53.3} & \multicolumn{1}{c}{79.6} & \multicolumn{1}{c}{75.7} & \multicolumn{1}{c}{85.2} & \multicolumn{1}{c|}{73.9} \\
    \hline
    R-SSD300*    & 07++12 & 76.4  & \multicolumn{1}{c}{88.0}  & \multicolumn{1}{c}{83.8} & \multicolumn{1}{c}{74.8} & \multicolumn{1}{c}{60.8} & \multicolumn{1}{c}{48.9} & \multicolumn{1}{c}{83.9} & \multicolumn{1}{c}{78.5} & \multicolumn{1}{c}{91.0} & \multicolumn{1}{c}{59.5} & \multicolumn{1}{c}{81.4} & \multicolumn{1}{c}{66.1} & \multicolumn{1}{c}{89.0} & \multicolumn{1}{c}{86.3} & \multicolumn{1}{c}{86.0} & \multicolumn{1}{c}{83.0} & \multicolumn{1}{c}{51.3} & \multicolumn{1}{c}{80.9} & \multicolumn{1}{c}{73.7} & \multicolumn{1}{c}{86.9} & \multicolumn{1}{c|}{73.8} \\
    \hline
    \end{tabular}%
    \end{tiny}%
    \caption{Results on VOC2012 test dataset using the networks trained by VOC07++12 train dataset (* is tested by ourselves)}
  \label{tab:addlabel23}
\end{table}%

Table \ref{tab:addlabel23} shows the results of VOC2012. All experiments were trained with VOC07++12 train dataset. 
YOLO \cite{redmon2016you} and YOLOv2 \cite{redmon2016yolo9000} have mAPs of 57.9\% and 73.4\%, respectively. SSD300 \cite{liu2016ssd} and DSSD321 \cite{fu2017dssd} show higher mAPs of 75.6\% and 76.5\%, respectively. Our R-SSD shows 76.6\% mAP, which is higher than those of SSD300 and DSSD321. R-SSD is also faster than DSSD321. 

  


\subsection{MS COCO example images by R-SSD}
\label{sec:image}

We test MS-COCO dataset and Fig.\ref{fig:ssdvspro} shows several resultant images among the dataset. R-SSD can prevent detecting multiple boxes for one object as shown in Fig.\ref{fig:ssdvspro}(b). Furthermore, it can efficiently increase the detection rate for  small objects as shown in Fig.\ref{fig:ssdvspro}(d).

\begin{figure}[!h]
\setlength{\abovecaptionskip}{1cm}
\begin{center}$
\begin{array}{cccc}
\includegraphics[width=2.8cm]{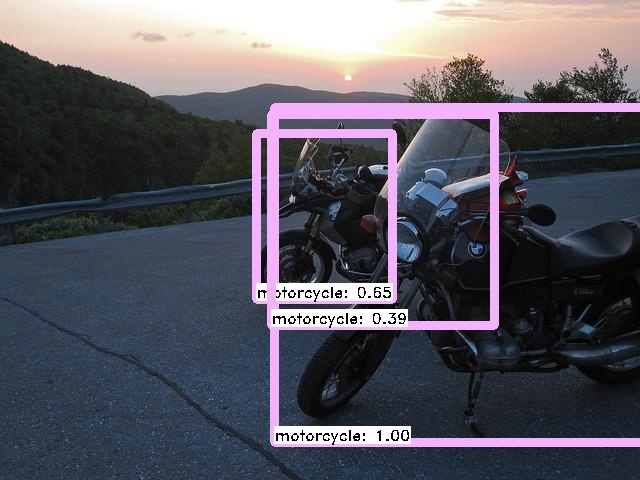}&
\includegraphics[width=2.8cm]{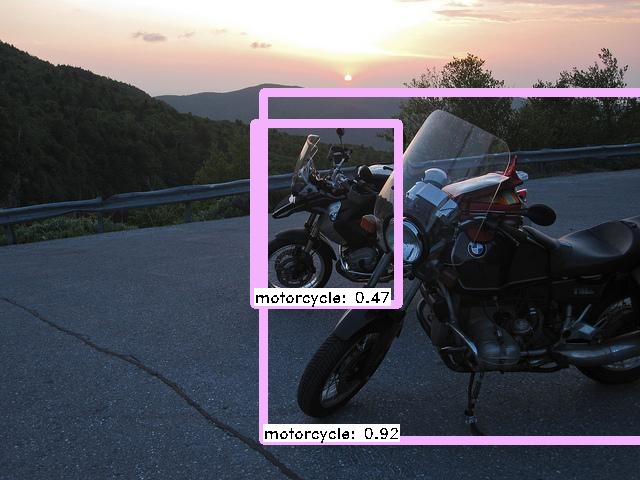}&
\includegraphics[width=2.8cm]{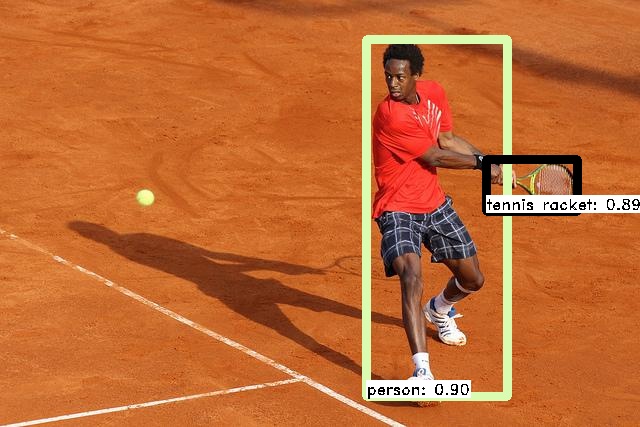}&
\includegraphics[width=2.8cm]{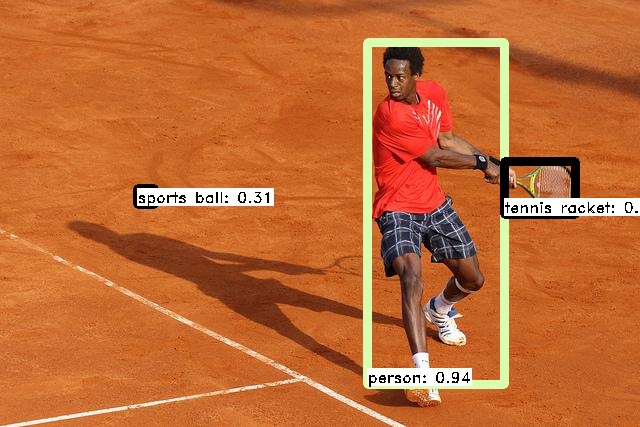}\\
\includegraphics[width=2.8cm]{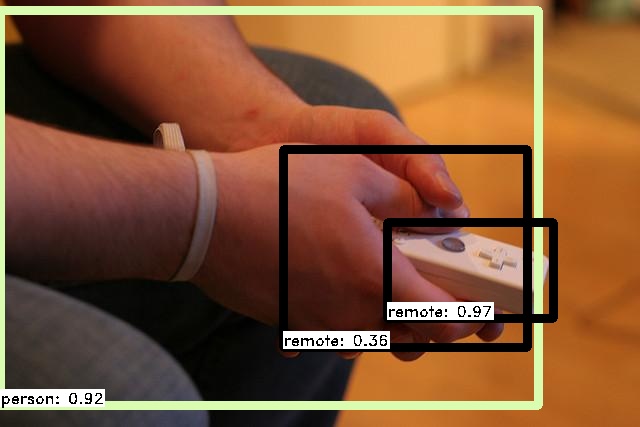}&
\includegraphics[width=2.8cm]{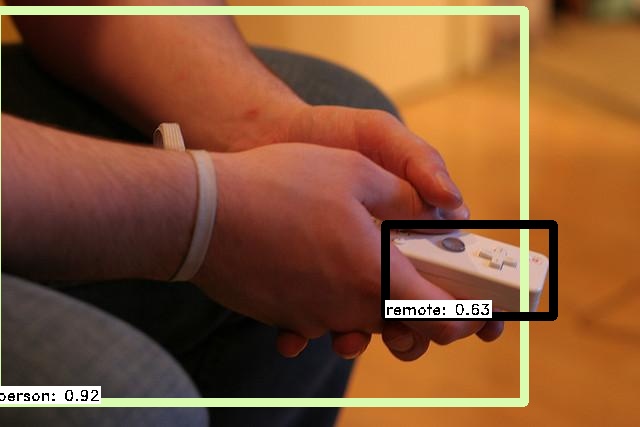}&
\includegraphics[width=2.8cm]{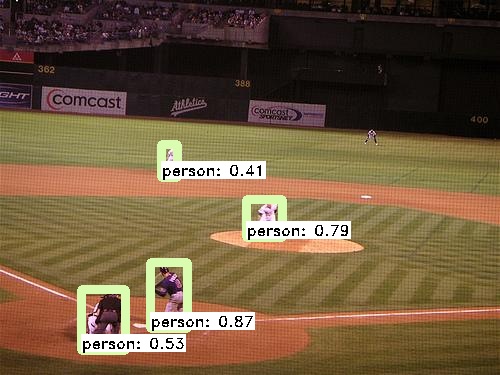}&
\includegraphics[width=2.8cm]{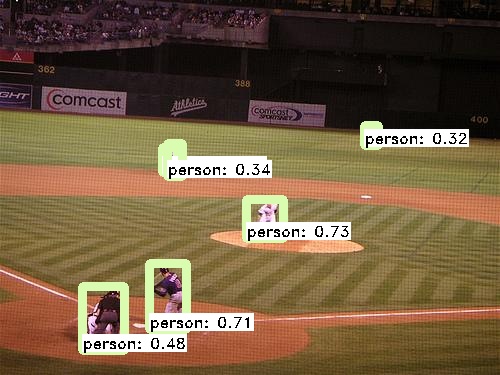}\\
\includegraphics[width=2.8cm]{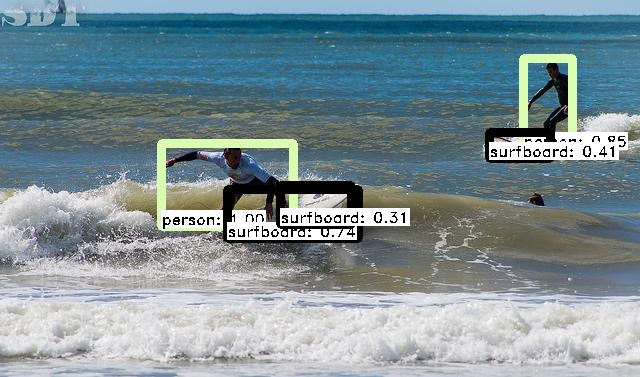}&
\includegraphics[width=2.8cm]{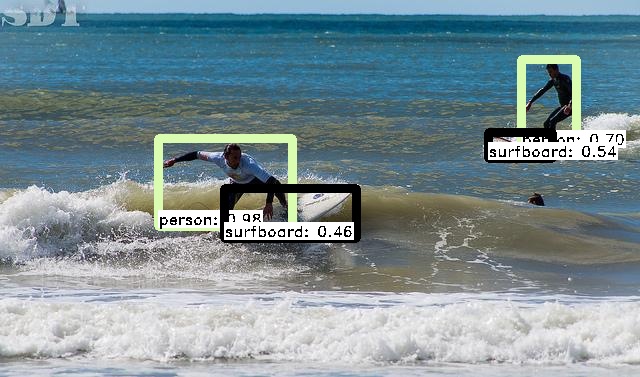}&
\includegraphics[width=2.8cm]{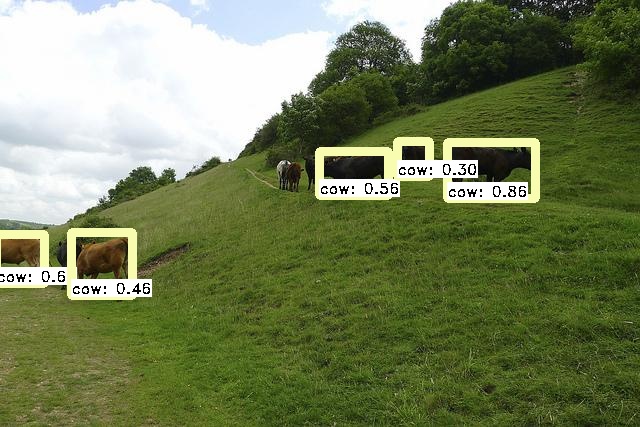}&
\includegraphics[width=2.8cm]{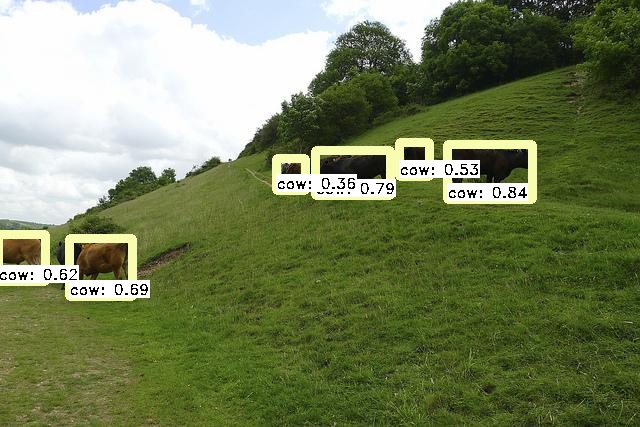}\\
\includegraphics[width=2.8cm]{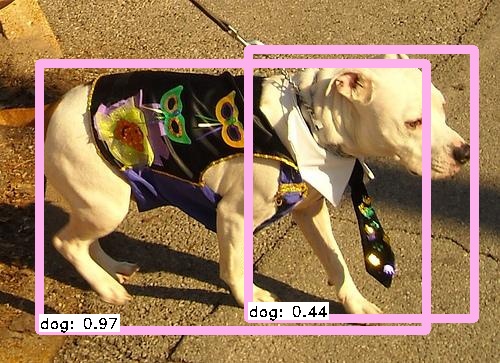}&
\includegraphics[width=2.8cm]{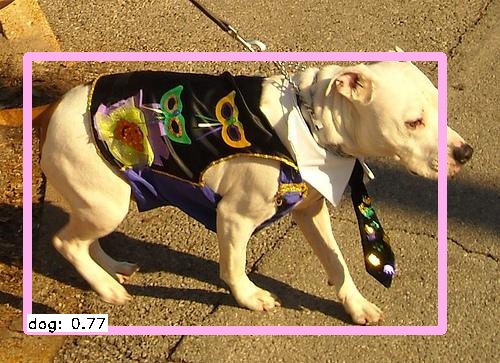}&
\includegraphics[width=2.8cm]{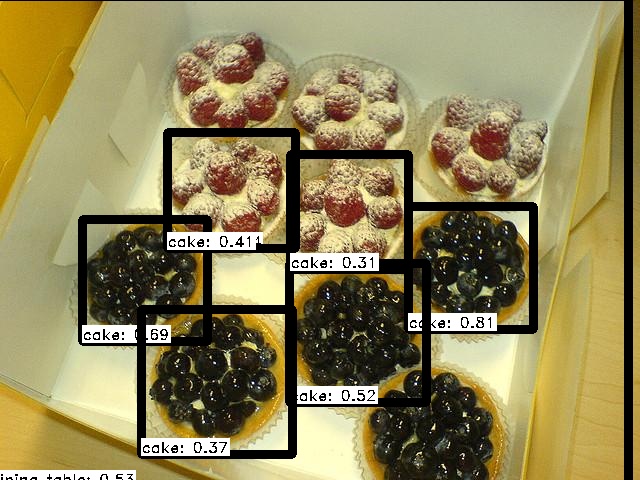}&
\includegraphics[width=2.8cm]{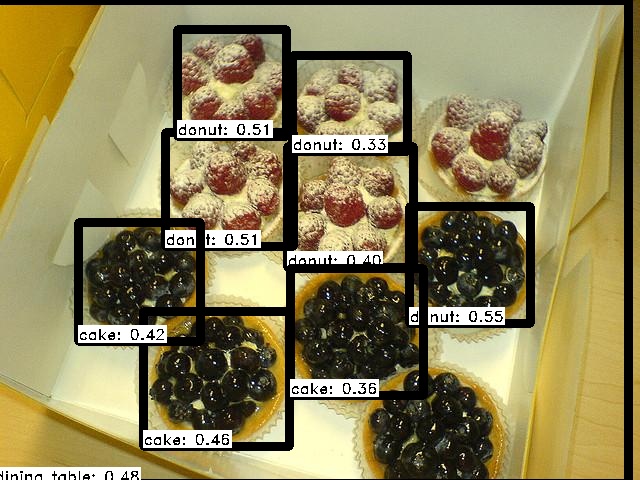}\\
(a)&(b)&(c)&(d)
\end{array}$
\end{center}
\caption{Conventional SSD vs. the proposed Rainbow SSD (R-SSD).  Boxes with objectness score of 0.3 or higher is drawn: (a) SSD with two boxes for one object; (b) R-SSD with one box for one object (c) SSD for small objects;  (d) R-SSD for small objects;}
\label{fig:ssdvspro}
\end{figure}



\bibliography{bmvc_review}

\end{document}